\pdfoutput=1

\documentclass[11pt,dvipsnames]{article}

\usepackage{ACL2023}

\usepackage{times}
\usepackage{latexsym}

\usepackage[T1]{fontenc}

\usepackage[utf8]{inputenc}

\usepackage{microtype}

\usepackage{inconsolata}

\usepackage{graphicx}
\graphicspath{ {./images/} }
\newcommand{\xlmr}{XLM\mbox{-}R}

\usepackage{tikz}
\usepackage{collcell}
\usepackage{multirow}

\makeatletter
\newcommand\R[1]{%
\ifdim#1pt<\z@ #1\else\phantom{-}#1\fi}
\makeatother

\newcommand*{\MinNumber}{0.70}%
\newcommand*{\MidNumber}{0.85} %
\newcommand*{\MaxNumber}{1.0}%

\newcommand{\C}[1]{%
    \hspace{-5pt}%
        \ifdim #1 pt > \MidNumber pt
            \pgfmathsetmacro{\PercentColor}{max(min(100.0*(#1 - \MidNumber)/(\MaxNumber-\MidNumber),100.0),0.00)} %
            \hspace{-0.33em}\colorbox{Green!\PercentColor!Yellow}{~#1~}
        \else
            \pgfmathsetmacro{\PercentColor}{max(min(100.0*(\MidNumber - #1)/(\MidNumber-\MinNumber),100.0),0.00)} %
            \hspace{-0.33em}\colorbox{Red!\PercentColor!Yellow}{~#1~}
        \fi\hspace{-11.3pt}\vspace{-0.2pt}
}

\usepackage{array}
\newcolumntype{P}[1]{>{\centering\arraybackslash}p{#1}}

\usepackage{arydshln} 

\title{Speaking Multiple Languages Affects the Moral Bias of Language Models}

\author{Katharina Hämmerl$^{1,2}$ \and Björn Deiseroth$^{3,4}$ \and Patrick Schramowski$^{4,5,9}$ \AND
Jindřich Libovický$^{6}$ \and Constantin A. Rothkopf$^{5,7,8}$ \AND Alexander Fraser$^{1,2}$ \and Kristian Kersting$^{4,5,8,9}$ \\
\\
$^1$Center for Information and Language Processing, LMU Munich, Germany \\
\texttt{\{lastname\}@cis.lmu.de}\\
$^2$Munich Centre for Machine Learning (MCML), Germany \\
$^3$Aleph Alpha GmbH, Heidelberg, Germany \\
$^4$Artificial Intelligence and Machine Learning Lab, TU Darmstadt, Germany \\
$^5$Hessian Center for Artificial Intelligence (hessian.AI), Darmstadt, Germany \\
$^6$Faculty of Mathematics and Physics, Charles University, Czech Republic \\
$^7$Institute of Psychology, TU Darmstadt, Germany \\
$^8$Centre for Cognitive Science, TU Darmstadt, Germany\\
$^9$German Center for Artificial Intelligence (DFKI) \\
}

\begin{document}
\maketitle
\begin{abstract}

Pre-trained multilingual language models (PMLMs) are commonly used when dealing with data from multiple languages and cross-lingual transfer.
However, PMLMs are trained
on varying amounts of data for each language.
In practice this means their performance is often much better on English than many other languages.
We explore to what extent this also applies to 
moral norms.
Do the models capture 
moral norms
from English and impose them on other languages?
Do the models exhibit random and thus potentially harmful beliefs in certain languages?
Both these issues could negatively impact cross-lingual transfer and potentially lead to harmful outcomes.
In this paper, we
(1) apply the \textsc{MoralDirection} framework to multilingual models, comparing results in German, Czech, Arabic, Chinese, and English, 
(2) analyse model behaviour on filtered parallel subtitles corpora, and 
(3) apply the models to a Moral Foundations Questionnaire, comparing with human responses from different countries.
Our experiments demonstrate that PMLMs do encode differing moral biases, but these do not necessarily correspond to cultural differences or commonalities in human opinions.
We release our code and models.\footnote{\url{https://github.com/kathyhaem/multiling-moral-bias}}

\end{abstract}

\section{Introduction}

\begin{figure}
\centering
\includegraphics[width=\linewidth]{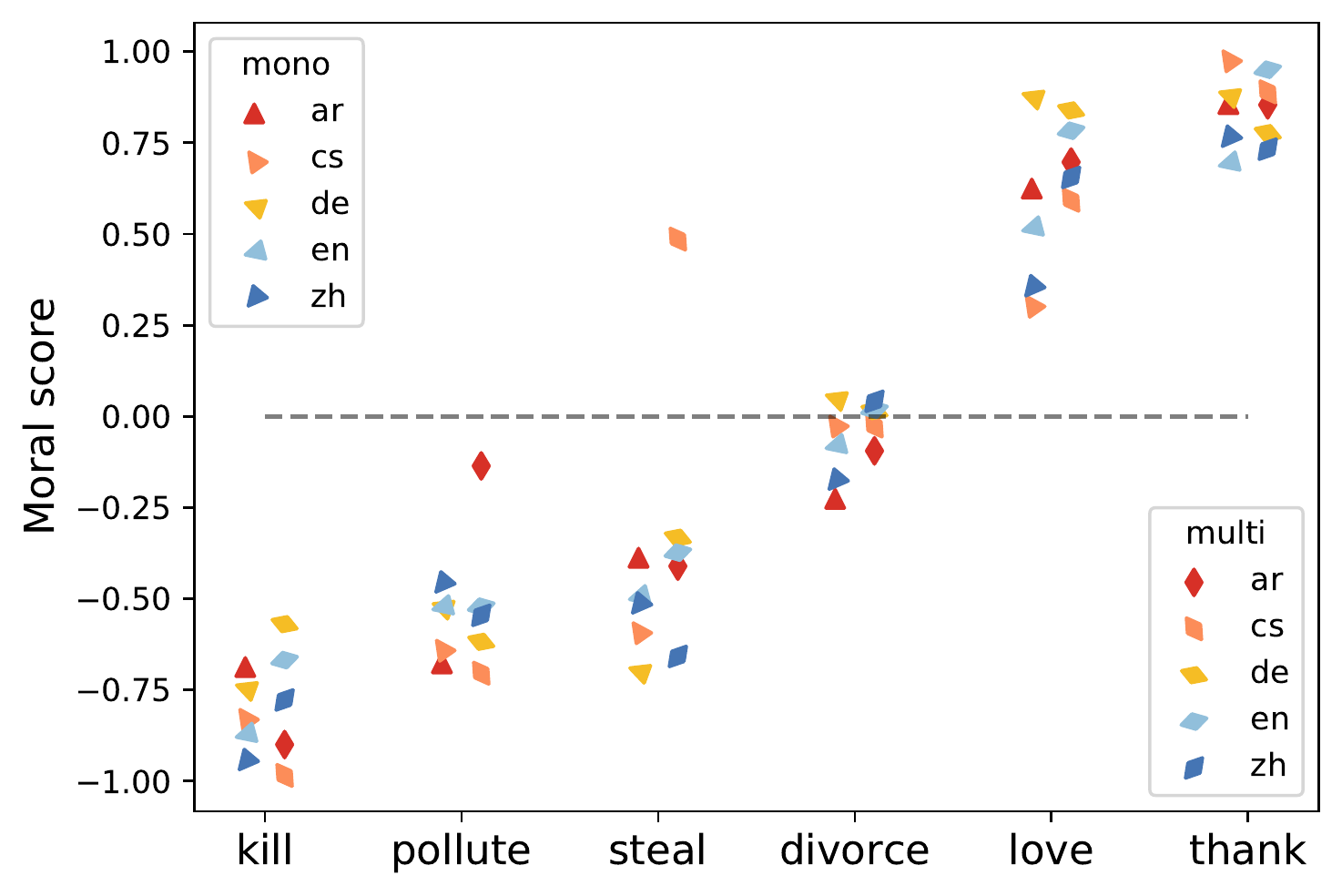}
\caption{\textsc{MoralDirection} score (y-axis) for several verbs (x-axis), as in \citet{schramowski22language}. We show scores for each language both from the respective monolingual model (triangles, left) and a multilingual model (rhombuses, right).
}
\label{fig:example-scores}
\end{figure}

Recent work demonstrated large pre-trained language models capture some symbolic, relational \citep{petroni-etal-2019-language}, but also commonsense \citep{davison-etal-2019-commonsense} knowledge.
The undesirable side of this property is seen in models reproducing biases and stereotypes (e.g., \citealp{caliskan-etal-2017-biases,choenni-etal-2021-stepmothers}).
However, in neutral terms, language models trained on large data from particular contexts will reflect cultural ``knowledge'' from those contexts.
We wonder
whether multilingual models will also reflect cultural knowledge from multiple contexts,
so we study moral intuitions and norms that the models might capture.

Recent studies investigated the extent to which language models reflect human values \citep{schramowski22language, fraser-etal-2022-moral}.
These works addressed monolingual English models.
Like them, we 
probe what the models encode, but we study multilingual models in comparison to monolingual models. 
Given constantly evolving social norms and differences between cultures and languages, we ask: 
Can a PMLM capture cultural differences, or does it impose a Western-centric view regardless of context?
This is 
a broad question which we cannot answer definitively.
However, we propose to analyse different aspects of mono- and multilingual model behaviour in order to come closer to an answer.
In this paper,
we pose three research questions, and present a series of experiments that address these questions qualitatively:

\begin{enumerate}
    \item If we apply the \textsc{MoralDirection} framework \citep{schramowski22language}
    to pretrained multilingual language models (PMLMs), how does this behave compared to monolingual models and to humans? (\S~\ref{sec:md-multiling})
    \item How does the framework behave when applied to parallel statements from a different data source? To this end, we analyse model behaviour on Czech-English and German-English OpenSubtitles data (\S~\ref{sec:opensub}).
    \item Can the mono- and multi-lingual models make similar inferences to humans on a Moral Foundations Questionnaire \citep{graham2011MFQ1}? Do they behave in ways that appropriately reflect cultural differences? (\S~\ref{sec:mfq})
\end{enumerate}

The three experiments reinforce each other in finding that our 
models grasp the moral dimension to some extent in all tested languages.
There are differences between the models in different languages, which sometimes
line up between multi- and mono-lingual models.
This does not necessarily correspond with differences in human judgements.
As an illustration, Figure~\ref{fig:example-scores} shows examples of the \textsc{MoralDirection} score for several verbs in our monolingual and multilingual models.

We also find that the models are very reliant on lexical cues, leading to problems like misunderstanding negation, and disambiguation failures.
This unfortunately makes it difficult to capture nuances.
In this work we compare the behaviour of the PMLM both to 
human data and to the behaviour of monolingual models in our target languages Arabic, Czech, German, Chinese, and English.

\section{Background}

\subsection{Pre-Trained Multilingual LMs}
PMLMs, such as \xlmr\ \citep{conneau-etal-2020-unsupervised}, are trained on large corpora of uncurated data, with an imbalanced proportion of language data included in the training.
Although sentences with the same semantics in different languages should theoretically have the same or similar embeddings,
this language neutrality is hard to achieve in practice \citep{libovicky-etal-2020-language}. 
Techniques for improving the model's internal semantic alignment (e.g., \citealp{zhao-etal-2021-inducing, cao2020multilingual,alqahtani-etal-2021-using-optimal, chi-etal-2021-improving, hammerl-etal-2022-combining}) have been developed, but these only partially mitigate the issue.
Here, we are interested in a more complex type of semantics and how well they are cross-lingually aligned.

\subsection{Cultural Differences in NLP}
Several recent studies deal with the question of how cultural differences affect NLP.
A recent comprehensive survey \citep{hershcovich-etal-2022-challenges} highlights challenges along the cultural axes of \textit{aboutness}, \textit{values}, \textit{linguistic form}, and \textit{common ground}.
Some years earlier, \citet{lin-etal-2018-mining} mined cross-cultural differences from Twitter data, focusing on named entities and slang terms from English and Chinese.
\citet{yin-etal-2022-geomlama} probed PMLMs for ``geo-diverse commonsense'', concluding that for this task, the models are not particularly biased towards knowledge about Western countries.
However, in their work the knowledge in question is often quite simple.
We are interested in whether this holds for more complex cultural values.
In the present study, we assume that using a country's primary language is the simplest way to probe for values from the target cultural context.
Our work analyses the extent to which one kind of cultural difference, moral norms, is
captured in PMLMs.

\subsection{Moral Norms in Pre-Trained LMs}
Multiple recent studies have investigated the extent to which language models reflect human values \citep{schramowski22language, fraser-etal-2022-moral}.
Further, benchmark datasets \citep{hendrycks21aligning,emelin-etal-2021-moral,ziems-etal-2022-moral} aiming to align machine values with human labelled data have been introduced.
Several such datasets \citep{forbes-etal-2020-social,hendrycks21aligning,alhassan-zhang-schlegel:2022:LREC} include scenarios from the ``Am I the Asshole?'' subreddit, an online community where users ask for an outside perspective on personal disagreements. 
Some datasets use the community judgements as labels directly, others involve crowdworkers in the dataset creation process.

Other works have trained models specifically to interpret moral scenarios, using such datasets.
A well-known example is \citet{jiang21delphi}, who propose a fine-tuned \textsc{Unicorn} model they call \textsc{Delphi}.
This work has drawn significant criticism, such as from
\citet{Talat2021AWO}, who argue ``that a model that generates moral judgments cannot avoid creating and reinforcing norms, i.e., being \textit{normative}''.
They further point out that the training sets 
sometimes
conflate moral questions with other issues such as medical advice or sentiments. 

\citet{hulpus-etal-2020-knowledge} explore a different direction. 
They project the Moral Foundations Dictionary,
a set of
lexical items related to foundations in Moral Foundations Theory (\S~\ref{subsec:mft}), onto a knowledge graph.
By scoring all entities in the graph for their relevance to moral foundations, they hope to
detect moral values expressed in a text.
\citet{solaiman-dennison-2021-palms} aim to adjust a pre-trained model to specific cultural values as defined in a targeted dataset.
For instance, they assert ``the model should oppose unhealthy beauty [...] standards''. 

A very interesting and largely unexplored area of research is to consider whether \textit{multilingual} language models capture differing moral norms.
For instance, moral norms in the Chinese space in a PMLM might systematically differ from those in the Czech space.
\citet{arora-etal-2022-probing} attempt to probe pre-trained models for cultural value differences using Hofstede's cultural dimensions theory \citep{hofstede1984culture} and the World Values Survey \citep{haerpfer2022world}.
They convert the survey questions to cloze-style question probes, obtaining score values by subtracting the output distribution logits for two possible completions from each other.
However, they find mostly very low correlations of model answers with human references.
Only a
few of their results show statistically significant correlations.
They conclude that the models differ between languages, but that these differences do not map well onto human cultural differences.

Due to the observation that the output distributions themselves do not reflect moral values well, we choose the \textsc{MoralDirection} framework for our studies.
In previous work, this approach identified a subspace of the model weights relating to a sense of ``right'' and ``wrong''
in English.

\subsection{Moral Foundations Theory}\label{subsec:mft}
Moral Foundations Theory \citep{HaidtJoseph2004MFT} is a comparative theory describing what it calls \textit{foundational moral principles}, whose relative importance can be measured to describe a given person's or culture's moral attitudes.
\citet{graham-etal-2009-liberals}
name the five factors ``Care/Harm'', ``Fairness/Reciprocity'', ``Authority/Respect'', ``Ingroup/Loyalty'', and ``Purity/Sanctity''.
Their importance varies both across international cultures \citep{graham2011MFQ1} and the (US-American) political spectrum \citep{graham-etal-2009-liberals}.
The theory has been criticised by some for its claim of innateness and its 
choice of factors, which has been described as ``contrived'' \citep{suhler-churchland-2011-critique}.
Nevertheless, the associated Moral Foundations Questionnaire \citep{graham2011MFQ1} has been translated into many languages and the theory used in many different studies (such as \citealp{joeckel-etal-2012-media, piazza-etal-2019-appraisals, DOGRUYOL2019109547}).
An updated version of the MFQ is being developed by \citet{atari_haidt_graham_koleva_stevens_dehghani_2022}.
In \S~\ref{sec:mfq}, we score these questions using our models and compare with human responses from previous studies on the MFQ.

\subsection{Sentence Transformers}\label{subsec:sentence-transformers}

By default, BERT-like models output embeddings at a subword-token level.
However, for many applications, including ours, sentence-level representations are necessary.
In our case, inducing the moral direction does not work well for mean-pooled token representations,
leading to near-random scores in many cases (see \S~\ref{subsec:md-framework}).
\citet{reimers-gurevych-2019-sentence} proposed 
Sentence-Transformers as a way to obtain meaningful, constant sized, sentence representations from BERT-like models.
The first Sentence-BERT (S-BERT) models were trained by tuning a pre-trained model on a sentence pair classification task.
By encoding each sentence separately and using a classification loss, the model learns 
more meaningful representations.

To obtain multilingual sentence representations, they proposed a student-teacher training approach using parallel corpora \citep{reimers-gurevych-2020-making},
where a monolingual S-BERT model acts as a teacher and a pre-trained multilingual model as a student model.
Such an approach forces the parallel sentences much closer together than in the original PMLM, which is not always desirable.
In our case, we might be unable to distinguish the effects of the S-BERT training from the original model,
which would interfere with probing the original model.

Unlike their work, 
we train a multilingual sentence transformer by translating the initial training data into our target languages (\S~\ref{subsec:our-sberts}),
and show that this is effective.
With this contribution, we show that multilingual S-BERT models can be trained in the same way as monolingual ones.
Our approach does not require a teacher-student training setup.
Note that we do require comparable datasets in size and ideally topics for each language.
While we do not explicitly align the data, we solve this by using machine translated versions of existing datasets, which means we have implicitly parallel data.

\section{Inducing Moral Dimensions in PMLMs}\label{sec:md-multiling}
We choose five languages to evaluate the behaviour of the multilingual models: Modern Standard Arabic (ar), Czech (cs), German (de), English (en), and Mandarin Chinese (zh).
These are all relatively high-resource languages, so we hope the model will be able to 
reliably
detect cultural knowledge in each language.
Since we rely on machine translation, using high-resource languages also ensures good translation quality.
We note here that languages and cultures or countries are at best approximately equivalent (cf. \citealp{lin-etal-2018-mining, sahlgren-etal-2021-basically}).
For instance, Arabic, English, and Chinese are standardised varieties that are written in a wide range of different contexts or cultures, which are likely
merged together
in the model to some degree.
However, separation by language is the best way we have to distinguish cultural contexts within a multilingual model.
As a point of comparison, we choose a monolingual language model for each of our target languages (see App.~\ref{app:models-used} for details).

\subsection{\textsc{MoralDirection} Framework}\label{subsec:md-framework}

We use the \textsc{MoralDirection} framework by \citet{schramowski22language}.
In the first step, this method encodes
a set of positively and negatively connotated action verbs with a sentence embedding model.
In \citet{schramowski22language}, this is an S-BERT model.
Each action verb is inserted into a set of ten template questions, such as ``Should I [verb]?'', ``Is it examplary to [verb]?'', and the output embedding for a verb is the mean over the embeddings of these questions.
Next, PCA is applied to the outputs, to obtain the ``moral direction'' subspace of the model.
Since the inputs are templates with only individual verbs changing, 
they are linguistically homogeneous, and the most salient differences for the PCA are the value judgements.
Ideally, a high amount of variance should be explained by the first principal component. 
The scores of these initial verbs are then normalised to lie within $[-1, 1]$. 
Subsequent scores can sometimes lie outside this range despite applying the normalisation.
The scores are then read as a value estimation along one axis, with scores around 0 being ``neutral'', scores close to -1 being very ``bad'', and scores close to 1 very ``good''.
However, note that the results we list in Tables~\ref{tab:initial-multiling-md}-\ref{tab:correlation_sbert_ours} and~\ref{tab:correlation_sbert_parallel} are correlations of model scores with user study data or correlations of model scores with other model scores.

We choose to use \textsc{MoralDirection} because it is able to work directly with sentence embeddings and extract a reasonably human-correlated moral direction from them, producing a value score along a single axis.
This makes it computationally inexpensive to transfer to other languages and datasets.
A drawback is that it is induced on short, unambiguous phrases, and can be expected to work better on such phrases.
Deriving a score along a single axis can also be limiting or inappropriate in certain contexts.
See also the discussion in Limitations.

For a list of the verbs and questions used to derive the transformation, see the source paper.
\citet{schramowski22language} also conduct 
a user study on Amazon MTurk 
to obtain reference scores for the statements in question. 

To test this method on multilingual and non-English monolingual models, we machine translate both the
verbs and the filled question templates used in the above study.
See Appendix~\ref{app:mt-quality} for the MT systems used, and a discussion of translation quality.
We edited some of the questions to ensure good translation.\footnote{e.g. ``smile to sb.'' $\to$ ``smile at sb.''}
Our primary measure is the correlation of resulting model scores with 
responses
from the
study in \citet{schramowski22language}.

We initially tested the method on mBERT \citep{devlin-etal-2019-bert} and \xlmr\ \citep{conneau-etal-2020-unsupervised}, as well as a selection of similarly sized monolingual models \citep{devlin-etal-2019-bert, antoun-etal-2020-arabert, straka-etal-2021-robeczech, chan-etal-2020-germans}, by mean-pooling their token representations.
See Appendix Table~\ref{tab:models-details} for a list of the models used.
Table~\ref{tab:initial-multiling-md} shows these initial results with mean-pooling.
\begin{table}[t]
\def\arraystretch{1.1}\tabcolsep=2.pt
\centering
\small
\begin{tabular}{lccccc}
\hline
\textbf{Model} & \textbf{en} & \textbf{ar} & \textbf{cs} & \textbf{de} & \textbf{zh} \\ 
\hline
mBERT (mean-pooled)       & \R{0.6}5 & \R{-0.10} & \R{0.12} & \R{-0.18} & \R{0.62} \\
XLM-R (mean-pooled)       & \R{-0.3}0 & \R{-0.07} & \R{-0.03} & \R{-0.14} & \R{0.10} \\
monolingual (mean-pooled) & \R{-0.1}3 & \R{0.46} & \R{0.07} & \R{0.10} & \R{0.70} \\
\hdashline
monolingual S-BERT-large & \R{0.79} &---&---&---&--- \\
XLM-R (S-BERT) & \R{0.85} & \R{0.82} & \R{0.85} & \R{0.83} & \R{0.81} \\
\hline
\end{tabular}
\caption{
Correlation of \textsc{MoralDirection} scores with user study data for different pre-trained mono- and multi-lingual models. 
First three rows used mean-pooled sentence embeddings; last two rows used embeddings resulting from sentence-transformers \cite{reimers-gurevych-2019-sentence}.
}
\label{tab:initial-multiling-md}
\end{table}
However, this generally did not achieve a correlation with the user study.
There were exceptions to this rule---i.e., the Chinese monolingual BERT, and the English and Chinese portions of mBERT.
This may be due to details in how the different models are trained, or how much training data is available for each language in the multilingual models.
Table~\ref{tab:initial-multiling-md} also includes results from the monolingual, large English S-BERT, and an existing S-BERT version of \xlmr\footnote{We used \texttt{sentence-transformers/xlm-r\-100langs-bert-base-nli-mean-tokens.}} \citep{reimers-gurevych-2020-making}.
These two models did show good correlation with the global user study,
highlighting that this goal requires semantic sentence representations.

\subsection{Sentence Representations}\label{subsec:our-sberts}
The existing S-BERT \xlmr\ model uses the student-teacher training with explicitly aligned data mentioned in \S~\ref{subsec:sentence-transformers}.
As we discuss there, we aim to change semantic alignment in the PMLM as little as possible before probing it. 
We also need S-BERT versions of the monolingual models.
Therefore, we train our own S-BERT models.
We use the sentence-transformers library \citep{reimers-gurevych-2019-sentence}, following their training procedure for training with NLI data.\footnote{{\url{https://github.com/UKPLab/sentence-transformers/blob/master/examples/training/nli/training_nli_v2.py}\nopagebreak}}
Although we do not need explicitly aligned data, we do require comparable corpora in all five languages, so we decide to use MNLI in all five languages.
In addition to the original English MultiNLI dataset \citep{williams-etal-2018-broad}, we take the German, Chinese and Arabic translations from XNLI \citep{conneau-etal-2018-xnli}, and provide our own Czech machine translations (cf. Appendix~\ref{app:mt-quality}).
Each monolingual model was tuned with the matching translation, while \xlmr$_{Base}$ was tuned with all five dataset translations.
Thus, our multilingual S-BERT model was not trained directly to align parallel sentences, but rather trained with similar data in each involved language
(without explicit alignment).
For more training details, see Appendix~\ref{app:sbert-training}.
We release the resulting S-BERT models to the Huggingface hub.

\subsection{Results}

Table~\ref{tab:multiling-md-sberts} shows the user study correlations of our S-BERT models.
\begin{table}[t]
\def\arraystretch{1.1}\tabcolsep=3.pt
\centering
\small
\begin{tabular}{lccccc}
\hline
\multirow{1}{*}{\begin{tabular}[l]{l}\textbf{Model}\end{tabular}} & \textbf{en} & \textbf{ar} & \textbf{cs} & \textbf{de} & \textbf{zh} \\ \hline
\multirow{2}{*}{\begin{tabular}[l]{l}XLM-R + MNLI \\ (S-BERT, all 5 langs)\end{tabular}}
 & \multirow{2}{*}{0.86} & \multirow{2}{*}{0.77} & \multirow{2}{*}{0.74} & \multirow{2}{*}{0.81} & \multirow{2}{*}{0.86} \\
 &&&&&\\
\multirow{2}{*}{\begin{tabular}[l]{l}monolingual + MNLI \\ (S-BERT, respective lang) \end{tabular}}
& \multirow{2}{*}{0.86} & \multirow{2}{*}{0.76} & \multirow{2}{*}{0.81} & \multirow{2}{*}{0.84} & \multirow{2}{*}{0.80} \\
&&&&&\\
\hline
\end{tabular}
\caption{
Correlation of \textsc{MoralDirection} scores from our mono- and multi-lingual S-BERT models with user study data.
}
\label{tab:multiling-md-sberts}
\end{table}
Clearly, sentence-level representations work much better for inducing the moral direction, and the method works similarly well across all target languages.
Figures~\ref{fig:example-scores} and~\ref{fig:more-scores} show examples of verb scores across models and languages, further illustrating that this method is a reasonable starting point for our experiments.

For Arabic and the Czech portion of \xlmr, the correlations are slightly lower than the other models.
Notably, Arabic and Czech are the smallest of our languages in \xlmr, at 5.4~GB and 4.4~GB of data \citep{wenzek-etal-2020-ccnet}, while their monolingual models contain 24~GB and 80~GB of data.

Since in the case of Czech, the correlation is higher in the monolingual model, and \xlmr\ and the monolingual model disagree somewhat (Table~\ref{tab:correlation_sbert_ours}), the lower correlation seems to point to a flaw of its representation in \xlmr.
For Arabic, the correlation of the monolingual model with English is similar to that seen in \xlmr, but the monolingual model also disagrees somewhat with the \xlmr\ representation (Table~\ref{tab:correlation_sbert_ours}).
This may mean there is actually some difference in attitude (based on the monolingual models), but \xlmr\ also does not capture it well (based on the \xlmr\ correlations).
Unfortunately, \citet{schramowski22language} collected no data specifically from Arabic or Czech speakers to illuminate this.

\begin{table}[tb]
\centering
\begin{tabular}{l|ccccc}
\textbf{language} &\textbf{en} & \textbf{ar}&\textbf{cs}&\textbf{de}&\textbf{zh} \\ \hline
\textbf{en} & \C{0.93} & \C{0.86} & \C{0.92} & \C{0.89} & \C{0.91} \\ 
\textbf{ar} & \C{0.86} & \C{0.84} & \C{0.89} & \C{0.89} & \C{0.86} \\ 
\textbf{cs} & \C{0.90} & \C{0.78} & \C{0.86} & \C{0.92} & \C{0.92} \\ 
\textbf{de} & \C{0.95} & \C{0.87} & \C{0.88} & \C{0.95} & \C{0.91} \\ 
\textbf{zh} & \C{0.94} & \C{0.89} & \C{0.84} & \C{0.94} & \C{0.94} \\ 
\end{tabular}
\caption{Correlation of languages between our S-BERT models on the user study questions. Below diagonal:
\xlmr\ model, tuned with MNLI data in five languages.
Above diagonal: Monolingual models, tuned with MNLI data in the respective languages.
On the diagonal: Correlation of the monolingual models with \xlmr\ in the respective language. 
}
\label{tab:correlation_sbert_ours}
\end{table}
In Table~\ref{tab:correlation_sbert_ours} we compare how much the scores correlate with each other when querying \xlmr\ and the monolingual models in different languages.
The diagonal shows correlations between the monolingual model of each language and \xlmr\ in that language.
Above the diagonal, we show how much the monolingual models agree with each other, while below the diagonal is the agreement of different languages within \xlmr.
On the diagonal, we compare each monolingual model with the matching language in \xlmr.
For English, German and Chinese, these show high correlations.
The lowest correlation overall is between the Czech and Arabic portions of \xlmr, while the respective monolingual models actually agree more.
The monolingual S-BERT models are generally at a similar level of correlation with each other as the multilingual model. 
German and Chinese, however, show a higher correlation with English in the multilingual model than in their respective monolingual models, which may show some interference from English.

We also show the correlations of languages within the pre-existing S-BERT model,\footnote{\texttt{sentence-transformers/xlm-r-100langs\-bert-base-nli-mean-tokens}} which was trained with parallel data, in Table~\ref{tab:correlation_sbert_parallel}.
Here, the correlations between languages are much higher, showing that parallel data training indeed changes the model behaviour on the \textit{moral dimension}.
These correlations are higher than that of any one model with the user study data, so this likely corresponds to an artificial similarity with English, essentially removing cultural differences from this model.

Summarised, the experiments in this section extend
\citet{schramowski22language} to a multilingual setting and indicate that multilingual LMs indeed capture moral norms. 
The high mutual correlations of scores show that the differences between models and languages are relatively small in this respect.
Note, however,
that the tested statements provided by \citet{schramowski22language} are not explicitly designed to grasp cultural differences.
We thus add further experiments
to address this question.

\section{Qualitative Analysis on Parallel Data}\label{sec:opensub}

\begin{table*}[htb]
\centering
\begin{tabular}{p{5.8cm}|p{4.4cm}|cccc}
& & \multicolumn{2}{c}{\textbf{monoling}} & \multicolumn{2}{c}{\textbf{XLM-R}} \\
\textbf{de} & \textbf{en} & \textbf{de} & \textbf{en} & \textbf{de} & \textbf{en} \\ 
\hline
Pures Gift. & Pure poison. & \R{-0.61} & \R{-0.71} & \textit{\R{0.65}} & \R{-0.69} \\
Ich erwürg dich! & I'll strangle you! & \R{-0.41} & \R{-0.58} & \textit{\R{0.90}} & \R{-0.62} \\
Hab jemandem einen Gefallen getan. & I did someone a favour. & \R{0.39} & \R{0.28} & \textit{\R{-0.41}} & \R{0.73} \\
Verräter ... wie Sie! & Traitors ... like you! & \R{-0.56} & \textit{\R{0.19}} & \R{-0.39} & \textit{\R{0.72}} \\
Sie brennen darauf, dich kennenzulernen. & They're dying to meet you. & \R{0.44} & \R{0.73} & \R{0.52} & \textit{\R{-0.31}} \\
Ich vermisse ihn sehr. & I really miss him. & \R{0.69} & \R{0.23} & \it \R{-0.41} & \it \R{-0.26} \\
Er schätzt mich. & He values me. & \R{1.12} & \R{0.31} & \textit{\R{0.04}} & \R{0.88} \\
\end{tabular}
\caption{Examples from the German-English OpenSubtitles data for which there is a large, spurious contrast between \textsc{MoralDirection} scores.
Scores that stand out as unreasonable are \textit{italicised}.
}
\label{tab:opensub-examples-de}
\end{table*}

To better understand how these models
generalise for various types of texts, we conduct a qualitative study using parallel data. 
For a parallel sentence pair, the \textsc{MoralDirection} scores should be similar in most cases.
Sentence pairs where the scores differ considerably may indicate cultural differences, or issues in the models.
In practice, very large score differences appear to be more related to the latter.
This type of understanding is important for further experiments with these models.

We conduct our analysis on OpenSubtitles parallel datasets
\citep{lison-tiedemann-2016-opensubtitles2016},\footnote{\url{http://www.opensubtitles.org/}}
which consist of relatively short sentences. 
Given that the \textsc{MoralDirection} is induced on short phrases, we believe that short sentences will be easier for the models.
The subtitles often concern people's behaviour towards each other, and thus may carry some moral sentiment. 
We use English-German and English-Czech data for our analysis.
To obtain the moral scores, we encode each sentence with the respective S-BERT model, apply the PCA transformation, and divide the first principal component by the normalising parameter.

Our analysis focuses on sentence pairs with very different scores.
We take steps to filter out mistranslated sentence pairs---see Appendix~\ref{app:filtering-details}.
Below, we discuss examples of where scores differ noticeably even when the translations are adequate.
Using Czech-English and German-English data, we compare the monolingual models with \xlmr, \xlmr\ with the monolingual models, and the monolingual models with each other. 
This analysis is based on manual inspection of 500 sentence pairs with the highest score differences for each combination.
Note that many of the sentence pairs were minor variations of each other, which significantly sped up the analysis.
Relevant examples are listed with their \textsc{MoralDirection} scores in Table~\ref{tab:opensub-examples-de} and Table~\ref{tab:opensub-examples-cs} in the Appendix.

\subsection{Reliance on Lexical Items}\label{ssec:lexical}

A common theme for many examples is an over-reliance on individual lexical items.
For example, ``Traitors ... like you!'' receives a positive score in English, while the German equivalent is correctly scored as negative.
Most likely, the English models took a shortcut: ``like you'' is seen as a good thing.

Similarly, \xlmr\ in English scores ``They're dying to meet you.'' somewhat negatively.
The English BERT gives a positive score. 
However, arguably this is a case where the most correct answer would be neutral, since this is more a positive sentiment than any moral concern.

\subsection{Multilinguality and Polysemy}

Continuing the theme of literalness, another dimension is added to this in the multilingual setting.
For instance, \xlmr\ scores the German ``Pures Gift.'' (\textit{pure poison}) as positive, likely because the key word ``Gift'' looks like English ``gift'', as in present.
However, the model also makes less explainable mistakes: many sentences with ``erwürgen'' (\textit{to strangle}) receive a highly positive score.

In the Czech-English data, there are even more obvious mistakes without a straightforward explanation.
Some Czech words are clearly not understood by \xlmr:
For instance, sentences with ``štědrý'' (\textit{generous}) are negative,
while any sentence with
``páčidlo'' (\textit{crowbar}) in it is very positive in \xlmr. 
Phrases with ``vrah'' (\textit{murderer}) get a positive score in XLM-R, possibly because of transliterations of the Russian 
word for medical doctor.
Most of these obvious mistakes of \xlmr\ are not present in RobeCzech.
However, ``Otrávils nás'' (\textit{You poisoned us}) receives a positive score from RobeCzech for unknown reasons.

Confusing one word for another can also be a problem within a single language: 
For example, ``Gefallen'' (a favour) receives a negative score from \xlmr\ in many sentences.
It is possible this model is confusing this with ``gefallen'' (past participle of ``fallen'', to fall), or some other similar word from a different language.
``Er schätzt mich'' and similar are highly positive in gBERT, as well as English \xlmr, but have a neutral score in German \xlmr.
Likely the latter is failing to disambiguate here, and preferring ``schätzen'' as in \textit{estimate}. 

\section{Moral Foundations Questionnaire}\label{sec:mfq}

\begin{figure*}
\centering
\includegraphics[width=\linewidth]{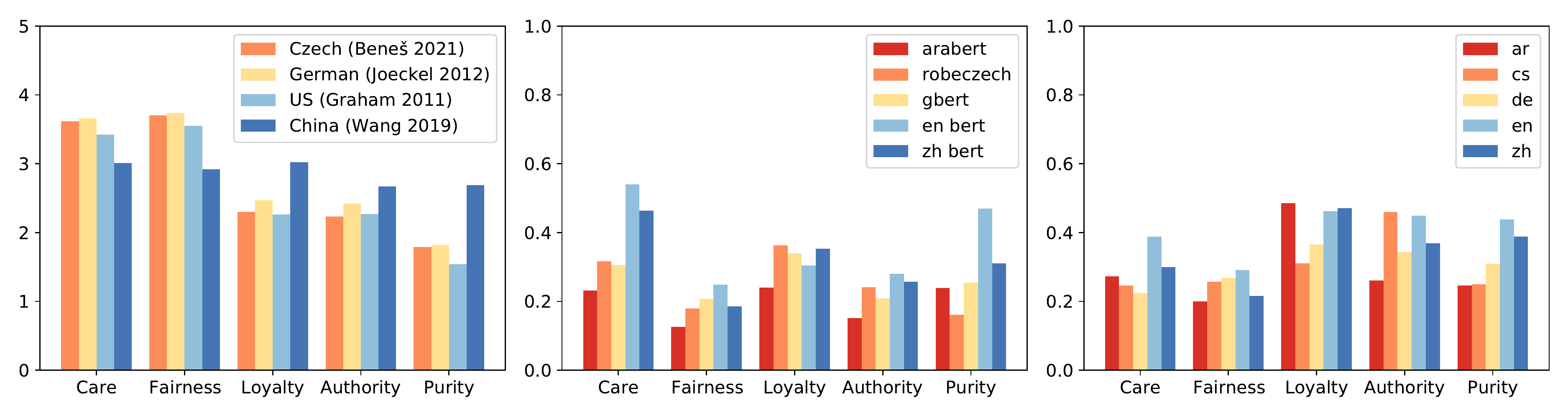}
\caption{MFQ aspect scores from humans and models. Left: Examples of human data from studies in different countries. Middle: Scores obtained from monolingual \textsc{MoralDirection} models. Right: Scores from \xlmr\ \textsc{MoralDirection} in five languages.
}
\label{fig:engineered}
\end{figure*}

\begin{figure}
\centering
\includegraphics[width=\linewidth]{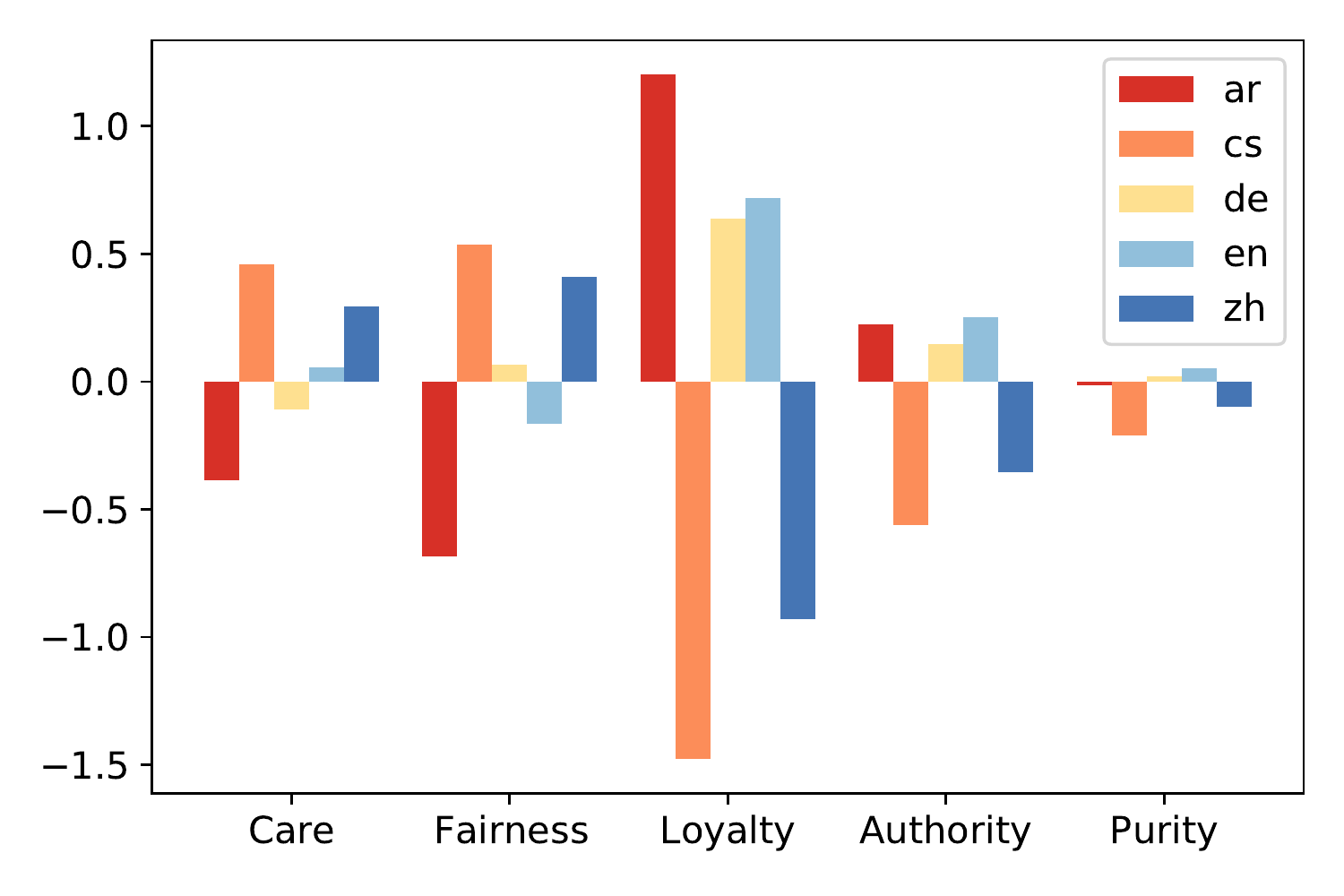}
\caption{Sanity check---MFQ aspect scores from the \xlmr\ \textsc{MoralDirection} models without Sentence-BERT tuning. This model had not obtained good correlations with human scores in \S~\ref{sec:md-multiling}.}
\label{fig:sanity-check}
\end{figure}

The MFQ has been applied in many different studies on culture and politics, meaning there is human response data from several countries available.
We pose the MFQ questions from \citet{graham2011MFQ1} to our models, in order to compare the model scores with data from previous studies.
We use the translations provided on the Moral Foundations website.\footnote{\url{https://moralfoundations.org/questionnaires/}}

Since the first part of the MFQ consists of very complex questions,
we rephrase these into simple
statements
(see Appendix~\ref{app:mfq-rephrased}).
Many of the statements in the first half of the questionnaire become \textit{reverse-coded} by simplifying them, that is, someone who values the aspect in question would be expected to answer in the negative.
For these statements, we multiply the model score by -1.
Further, we know that language models struggle with negation \citep{kassner-schutze-2020-negated}, so we remove ``not'' or ``never'' from two statements
and flip the sign accordingly.
In the same way, we remove ``a lack of'' from two statements.

These adjustments already improved the coherence of the resulting aspect scores, 
but we found further questions being scored by the models as if reverse-coded, i.e., with a negative score when some degree of agreement was expected.
These were not simply negated statements, but they did tend to contain lexical items that were strongly negatively associated, and in multiple cases contained a negative moral judgement of the action or circumstance in question.
Because the models appear to be so lexically focused (see \S~\ref{ssec:lexical}), this combination led to a strong negative score for some of these questions.
We decided to rephrase such statements as well, usually flipping their sign while changing the wording as little as possible.
Still, we note here that this should be considered a type of prompt engineering, and that implicatures of the statements may have changed through this process.
We provide the list of rephrased English statements and multipliers in Appendix Table~\ref{tab:mfq_prompt-engineered}.

We manually apply the same changes to the translations.
The full list of English and translated statements, as well as model scores for each question, is available as a CSV file.
Finally, we mean-pool the question scores within each aspect to obtain the aspect scores.
Most of the model scores for each question will be within [-1, 1].
The results are shown in Figure~\ref{fig:engineered}.

\subsection{Human Response Data}

Also in Figure~\ref{fig:engineered}, we show German data from \citet{joeckel-etal-2012-media}, Czech data from \citet{Benes2021thesis}, US data from \citet{graham2011MFQ1}, and Chinese data from \citet{wang-etal-2019-homophobia}
for comparison.
Note that these are not necessarily representative surveys. 
The  majority of the data in question were collected primarily in a university context and the samples skew highly educated and politically left.
For Germany, the US and the Czech Republic, the individual variation, or variation between political ideologies seems to be larger than the variation between the countries.
The
Chinese sample scores more similarly to conservative respondents in the Western countries.
Although many individuals score in similar patterns as the average, the difference between individuals in one country can be considerable.
As an example, see Figure~\ref{fig:human-var-example}. 

None of our models' scores map directly onto average human responses.
The model scores do not use the full range of possible values, but even the patterns of relative importance do not match the average human patterns.
Scores sometimes vary considerably in different models and different languages within \xlmr, and not necessarily in a way that would follow from cultural differences.
The average scores within \xlmr\ are somewhat more similar to each other than the scores from the monolingual models are, giving some weak evidence that the languages in the multilingual model assimilate to one another.
However, some differences between the monolingual models are also reflected in the multilingual model.

\subsection{Sanity Check}

We compare against scores from the unmodified, mean-pooled \xlmr\ models, shown in Figure~\ref{fig:sanity-check}.
These models did not have the Sentence-BERT tuning applied to them, but otherwise we used the same procedure to obtain the scores.
The inconsistent and very unlike human scores reinforce the finding from \S~\ref{sec:md-multiling} that mean-pooled representations are not useful for our experiments.
They also confirm that the results in our main MFQ experiments are not arbitrary.

\section{Conclusions}

We investigated the \textit{moral dimension} of pre-trained language models in a multilingual context.
In this section, we 
discuss our research questions:

\paragraph{(1) Multilingual \textsc{MoralDirection}.}
We applied the \textsc{MoralDirection} framework to \xlmr, as well as monolingual language models in five languages.
We were able to induce models that correlate with human data similarly well as their English counterparts in \citet{schramowski22language}.
We analysed differences and similarities across languages.

In the process, we showed that sentence-level representations, rather than mean-pooled token-level representations, are necessary in order to induce a reasonable moral dimension for most of these models.
We trained monolingual S-BERT models for our five target languages Arabic, Czech, German, English, and Mandarin Chinese.
As well, we created a multilingual S-BERT model from \xlmr\ which was trained with MNLI data in all five target languages.

\paragraph{(2) Behaviour on Parallel Subtitles.}
A limitation of the \textsc{MoralDirection} is that it is induced on individual words, and thus longer sentences are a significant challenge for the models.
Still, we were able to test them on parallel subtitles data, which contains slightly longer, but predominantly still short, sentences.
Problems that showed up repeatedly in this experiment were an over-reliance on key lexical items and a failure to understand compositional phrases, particularly negation.
Additionally, typical problems of PMLMs, such as disambiguation problems across multiple languages, were noticeable within \xlmr.
Non-English languages appeared more affected by such issues, despite the fact that all our target languages are relatively 
high resource.

\paragraph{(3) Moral Foundations Questionnaire.}
Our experiments with the MFQ reinforce the conclusion that the \textsc{MoralDirection} models capture a general sense of right and wrong, but do not display entirely coherent behaviour.
Again, compositional phrases and negation were an issue in multiple cases.
We had set out to investigate whether cultural differences are adequately reflected in the models' cross-lingual behaviour.
However, our findings indicate that rather, there are other issues with the cross-lingual transfer that mean we cannot make such nuanced statements about the model behaviour.
To the extent that model behaviour differs for translated data, this does not seem to match cultural differences between average human responses from different countries. 

We had initially wondered whether models would impose values from an English-speaking context on other languages.
Based on this evidence,
it seems that the models do differentiate between cultures to some extent, but there are caveats:
The differences are not necessarily consistent with human value differences, which means the models are not always adequate.
The problem appears to be worse when models are trained on smaller data for a given language.
Meanwhile, German and Chinese have noticeably high agreement with English in our multilingual model, and all languages are extremely highly correlated in the pre-existing parallel-data S-BERT model (Table~\ref{tab:correlation_sbert_parallel}).
This clearly shows that training with parallel data leads to more similar behaviour in this dimension, more or less removing cultural differences, but indeed there may be some transference even without parallel data.

\paragraph{Future Work.}
This leads to several future research questions: 
(i) Can we reliably investigate encoded (moral) knowledge reflected by PMLMs on latent representations or neuron activations? 
Or do we need novel approaches? 
For instance, \citet{jiang21delphi} suggest evaluating the output of generative models and, subsequently, \citet{arora-etal-2022-probing} apply masked generation using PMLMs to probe cultural differences in values. 
However, the generation process of LMs is highly dependent, among other things, on the sampling process. 
Therefore, it is questionable if such approaches provide the required transparency. 
Nevertheless, \citet{arora-etal-2022-probing} come to a similar conclusion as indicated by our results: 
PMLMs encode differences between cultures. 
However, these are weakly correlated with human surveys, which leads us to the second future research question:
(ii) How can we reliably teach large-scale LMs to reflect cultural differences but also commonalities? Investigating PMLMs'
moral direction and probing the generation process leads to inconclusive results, i.e., these models encode differences, which, however, do not correlate with human opinions. 
But correlating with human opinions
is a requirement for models to work faithfully in a cross-cultural context. Therefore, we advocate 
for
further research on teaching cultural characteristics to LMs.

\section*{Limitations}
The \textsc{MoralDirection} framework works primarily for short, unambiguous phrases.
While we show that it is somewhat robust to longer phrases, it does not deal well with negation or certain types of compositional phrases.
We showed that in such cases, prompt engineering seems to be necessary in order to get coherent answers.
Inducing the \textsc{MoralDirection} was done on a small set of verbs, and the test scenarios in this paper---apart from \S~\ref{sec:opensub}---are also relatively small.

The scope of our work is specific to our stated target languages, which are all relatively high-resource, meaning the method may not hold up for languages with smaller corpora, especially in the context of PMLMs.
This work presents primarily an exploratory analysis and qualitative insights.

Another point is that the monolingual models we used may not be precisely comparable.
Table~\ref{tab:models-details} lists details of parameter size, training, tokenizers, data size and data domain.
The models are all similarly sized, but data size varies considerably.
\xlmr\ and RobeCzech do not use next sentence prediction as part of their training objective.
However, the authors of RoBERTa \citep{liu-etal-2019-roberta} argue this difference does not affect representation quality.
Further, exactly comparable models do not exist for every language we use.
We rather choose well-performing, commonly-used models.
Thus, we believe the model differences play a negligible role in the context of our scope.

More broadly speaking, the present work makes the strong assumption that cultural context and language are more or less equivalent, which does not hold up in practice.
Furthermore, \textsc{MoralDirection}, like related methods, only consider a single axis, representing a simplistic model of morality.
In the same vein, these models will output a score for any input sentence, including morally neutral ones, sometimes leading to random answers.

\section*{Broader Impacts}
Language models should not decide moral questions in the real world, but research in that direction might suggest that this is in fact possible.
Besides undue anthropomorphising of language models, using them to score moral questions could lead to multiple types of issues:
The models may reproduce and reinforce questionable moral beliefs.
The models may hallucinate beliefs. 
And particularly in the context of cross-lingual and cross-cultural work, humans might base false, overgeneralising, or stereotyping assumptions about other cultures on the output of the models.

\section*{Acknowledgements}

We thank Sven Jöckel for providing us with their raw results from their MFQ 
studies, and Hashem Sellat and Wen Lai for their help with formulating
the MFQ questions in Arabic and Chinese.
Thank you to Morteza Dehghani. 

This publication was supported by LMUexcellent, funded by the Federal Ministry of Education and Research (BMBF) and the Free State of Bavaria under the Excellence Strategy of the Federal Government and the Länder; and by the German Research Foundation (DFG; grant FR 2829/4-1).
The work at CUNI was supported by Charles University project PRIMUS/23/SCI/023 and by the European Commission via its Horizon research and innovation programme (No. 870930 and 101070350).
Further, we gratefully acknowledge support by the Federal Ministry of Education and Research (BMBF) under Grant No. 01IS22091.
This work also benefited from the ICT-48 Network of AI Research Excellence Center “TAILOR" (EU Horizon 2020, GA No 952215), the Hessian research priority program LOEWE within the project WhiteBox, the Hessian Ministry of Higher Education, and the Research and the Arts (HMWK) cluster projects “The Adaptive Mind” and “The Third Wave of AI”.

\bibliography{anthology,custom}
\bibliographystyle{acl_natbib}

\appendix

\section{Details of Models Used}\label{app:models-used}

Table~\ref{tab:models-details} lists the models we tuned and evaluated with their exact names, sizes, objectives and data.

The models are all of a similar size, although data size varies by up to one order of magnitude between monolingual models.
\xlmr\ has much larger data in total, but data size for individual languages is more comparable to the other models.
The data domains vary but overlap (Web, Wiki, News).
\xlmr\ and RobeCzech do not use next sentence prediction as part of their training objective.
However, we believe these differences play a negligible role in the context of our work.

\begin{table*}[tb]
\centering\footnotesize

\begin{tabular}{cP{5cm}ccccP{2cm}}
\hline
\textbf{Lng} & \textbf{Name} & \textbf{Params} & \textbf{Objective} & \textbf{Tokenizer} & \textbf{Data size} & \textbf{Domain} \\ \hline

ar & aubmindlab/bert-base-arabertv02 \citep{antoun-etal-2020-arabert} & 110M & MLM+NSP  & SP, 60k & 24~GB & Wiki, News \\

cs & ufal/robeczech-base \citep{straka-etal-2021-robeczech} & 125M & MLM & BPE, 52k & 80~GB & News, Wiki, Web \\
de & deepset/gbert-base \citep{chan-etal-2020-germans} & 110M & MLM+NSP & WP, 31k & 136~GB & Web, Wiki, Legal \\
en & bert-base-cased \citep{devlin-etal-2019-bert} & 110M & MLM+NSP  & WP, 30k & 16~GB & Books, Wiki \\
zh & bert-base-chinese \citep{devlin-etal-2019-bert} & 110M & MLM+NSP & WP, 21k & ? & Wiki \\ \hline

--- & xlm-roberta-base \citep{conneau-etal-2020-unsupervised} & 125M & MLM & SP, 250k  & 2.5TB & Web \\

\hline
\end{tabular}

\caption{The monolingual pre-trained language models used. 
We tuned each model with the S-BERT framework before using it for our experiments. 
Objectives: MLM = masked language modelling, NSP = next sentence prediction, Tokenization: WP = WordPiece, SP = SentecePiece, unigram model.}
\label{tab:models-details}
\end{table*}

\section{Machine Translation Quality}\label{app:mt-quality}

Machine translation is used to translate the templated sentences from English into 
Arabic, Czech, German and Chinese.
For Arabic and Chinese, we use Google Translate.
The sentences are short and grammatically very simple.

For translation into Czech, we use CUBBITT \citep{popel2020transforming}, a machine translation system that scored
in the first cluster in WMT evaluation campaigns 2019--2021.
For translation into German, we use the WMT21 submission of the University of Edinburgh \citep{chen-etal-2021-university}.
To validate our choice of machine translation systems, we estimate the translation quality using
the reference-free version of the COMET score \citep{rei-etal-2020-comet} (model \texttt{wmt21-comet-qe-mqm}) on the 2.7k generated questions.

\begin{table}
    \centering
    \centering\footnotesize
    \begin{tabular}{llc}
         \hline
         \textbf{Lng} & \textbf{Model} & \textbf{COMET} \\ \hline
         
         ar & Google Translate & .1163 \\
            & \textcolor{gray}{OPUS MT 2022}     & \textcolor{gray}{.1183} \\
            & \textcolor{gray}{OPUS MT 2020}     & \textcolor{gray}{.1163} \\
         \hline
         cs & CUBBITT & .1212 \\
            & \textcolor{gray}{OPUS MT 2022}     & \textcolor{gray}{.1212} \\
            & \textcolor{gray}{OPUS MT 2020}     & \textcolor{gray}{.1197} \\
            & \textcolor{gray}{Google Translate} & \textcolor{gray}{.1193} \\
         \hline
         de & UEdin WMT21      & .1191 \\
            & \textcolor{gray}{Facebook, WMT19}  & \textcolor{gray}{.1191} \\
            & \textcolor{gray}{Google Translate} & \textcolor{gray}{.1190} \\
            & \textcolor{gray}{OPUS MT 2022}     & \textcolor{gray}{.1180} \\
            & \textcolor{gray}{OPUS MT 2020}     & \textcolor{gray}{.1123} \\
         \hline
         zh & Google Translate & .1111 \\ 
            & \textcolor{gray}{OPUS MT 2020}     & \textcolor{gray}{.1101} \\
         \hline
    \end{tabular}
    \caption{Machine translation quality of the templated sentences use in the MoralDimension estimation measured by the reference-free COMET score. Unused alternatives are in gray.}
    \label{tab:moral-comet}
\end{table}

To train the S-BERT models, we use the TRANSLATE-TRAIN part of the XNLI dataset
that is distributed with the dataset (without specifying what translation system
was used).
For translation into Czech, we use CUBBITT again.
To ensure the 
translation quality is comparable, we use the same evaluation metric as in the previous case on 5k randomly sampled sentences.

\begin{table}
    \centering\footnotesize
    \begin{tabular}{llc}
         \hline
         \textbf{Lng} & \textbf{Model} & \textbf{COMET} \\ \hline
         ar & \multirow{3}{*}{as in XNLI} & .1013 \\
         de & & .1051 \\
         zh & & .1017 \\ \hline
         cs & CUBBITT & .1153 \\
         & \textcolor{gray}{Google Translate} & \textcolor{gray}{.1150} \\
         & \textcolor{gray}{OPUS MT 2022} & \textcolor{gray}{.1144} \\
         & \textcolor{gray}{OPUS MT 2020} & \textcolor{gray}{.1126} \\
         \hline
    \end{tabular}
    \caption{Machine translation quality of the MNLI data used for training S-BERT models measured by the reference-free COMET score. Unused alternatives are in gray.}
    \label{tab:nli-comet}
\end{table}

\section{Correlations in Existing (Parallel Data) S-BERT}\label{app:corr-parallel}

Table~\ref{tab:correlation_sbert_parallel} shows the correlations of languages within the pre-existing multilingual S-BERT, trained with parallel data.
The correlations within this model are extremely high, considerably higher than that of any one model with the user study.

\begin{table}[htb]
\centering
\begin{tabular}{l|ccccc}
\textbf{language} &\textbf{en} & \textbf{ar}&\textbf{cs}&\textbf{de}&\textbf{zh} \\ \hline
\textbf{en} &  \\
\textbf{ar} & \C{0.97} &  \\
\textbf{cs} & \C{0.98} & \C{0.97} &  \\
\textbf{de} & \C{0.98} & \C{0.97} & \C{0.98} &  \\
\textbf{zh} & \C{0.97} & \C{0.96} & \C{0.96} & \C{0.96} &  \\
\end{tabular}
\caption{In-model correlation of scores on the user study questions, within \texttt{sentence-transformers/xlm-r-
100langs-bert-base-nli-mean-tokens}.
}
\label{tab:correlation_sbert_parallel}
\end{table}

\section{Sentence-BERT Tuning Procedure}\label{app:sbert-training}

We follow the training script provided by \citet{reimers-gurevych-2019-sentence} in the sentence-tranformers repository.
As training data, we use the complete MNLI (\citealp{williams-etal-2018-broad}; 433k examples) in the five respective languages.
The dev split from the STS benchmark (\citealp{cer-etal-2017-semeval}; 1500 examples) serves as development data.
We also machine translate this into the target languages.
The loss function is Multiple Negatives Ranking Loss \citep{henderson-etal-2017-multiple-negatives}, which benefits from larger batch sizes.
We use sentence-transformers version 2.2.0.
Table~\ref{tab:sbert-params} lists further training parameters.

\begin{table}
\centering\footnotesize

\begin{tabular}{cc}
\hline
    \bf Parameter & \bf Value \\
    \hline
    Batch size & 128 \\
    Max seq length & 75 \\
    Epochs & 1 \\
    Warmup & 10\% of train data \\
    Save steps & 500 \\
    Optimizer & AdamW \\
    Weight decay & 0.01 \\
    \hline
\end{tabular}
\caption{Sentence-BERT tuning parameters.}\label{tab:sbert-params}
\end{table}

\section{Computational Resources}

In addition to the six models used for further experiments, we trained five \xlmr\ with single-language portions of data.
Each of the monolingual models, as well as the \xlmr\ versions tuned with one part of the data, took around 0.6 hours to train.
Tuning \xlmr\ with data in all five languages accordingly took around three hours.
S-BERT tuning was done on one Tesla V100-SXM3 GPU, with 32 GB RAM, at a time.
We also trained one version of \xlmr\ on English data with a smaller batch size on an NVIDIA GeForce GTX 1080 GPU with 12 GB RAM.
In all other experiments, the language models were used in inference mode only, and they were mostly run on the CPU.

\section{Variance in \textsc{MoralDirection} Scores}

In this section we discuss another aspect of \textsc{MoralDirection} scores in multilingual versus monolingual models:
How much they vary between different languages for each statement.
For instance, if the variance is smaller in the multilingual model, this would mean that the multilingual model applies more similar judgements across languages.
To quantify this, we calculate the score variance for each of the basic verbs from ~\citet{schramowski22language} over the five monolingual models, as well as over the five portions of the multilingual model.

\begin{figure}
\centering
\includegraphics[width=\linewidth]{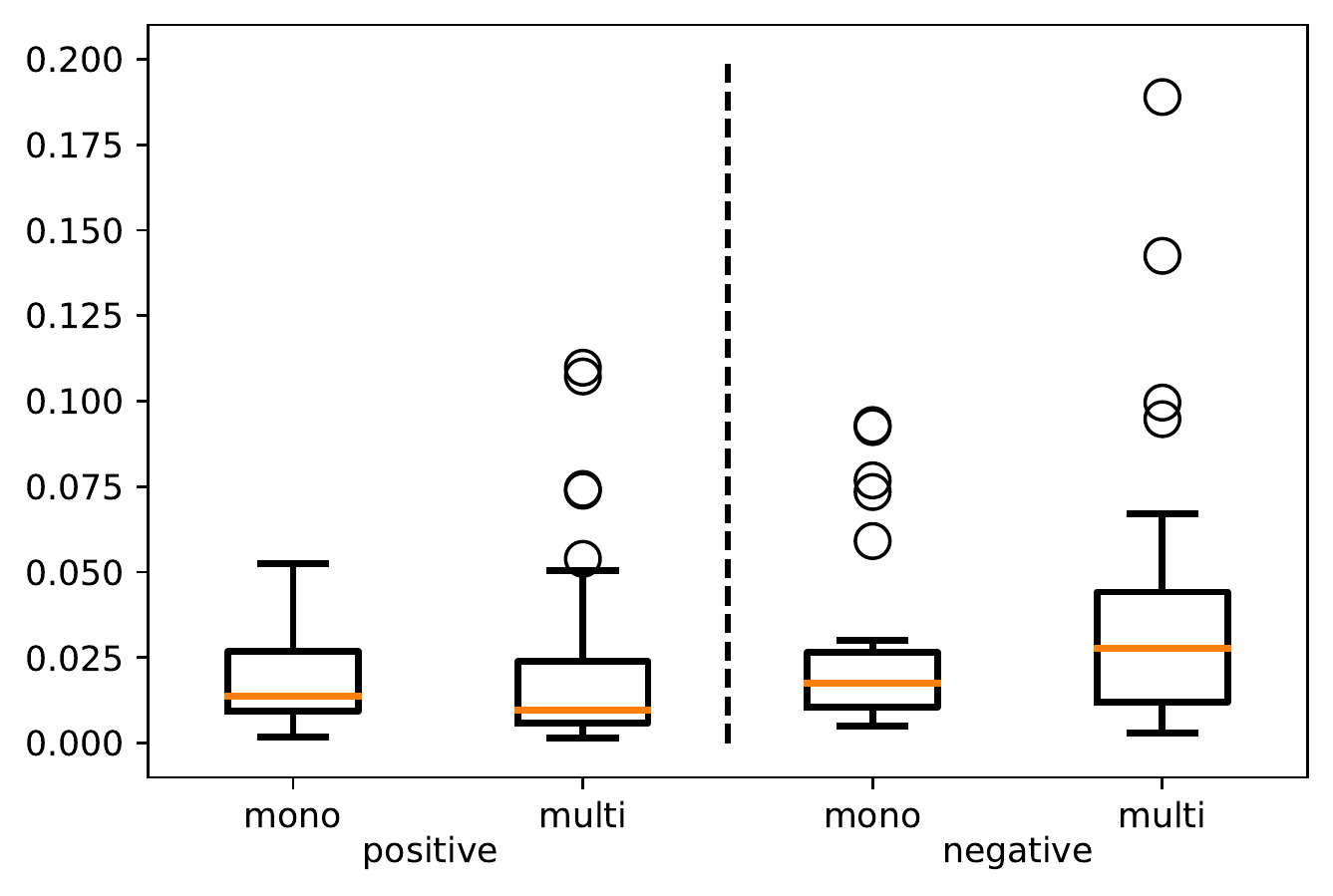}
\caption{\textsc{MoralDirection} score variances throughout all languages of the verbs taken from~\citet{schramowski22language}, on monolingual and multilingual models. The verbs are further grouped into positive and negative based on mean scores.}
\label{fig:boxplot-variance}
\end{figure}

We furthermore grouped the verbs into ``positive'' and ``negative'', depending on whether their mean score from the multilingual model is greater or lower than zero.
This results in 35 positive and 29 negative verbs.
Figure~\ref{fig:boxplot-variance} shows box-plots of the variance for those groups.
Overall, variances are similar for monolingual and multilingual models.
The positive verbs have a lower variance in the multilingual than in the monolingual models.
However, the opposite is true for the group of negative verbs, averaging out to very similar variances overall.
Therefore, analysing variances does not lead us to conclusions about differing behaviour of monolingual versus multilingual models.

\section{More Examples \textsc{MoralDimension} for Verbs}

\begin{figure*}
    \centering
    \includegraphics[width=\textwidth]{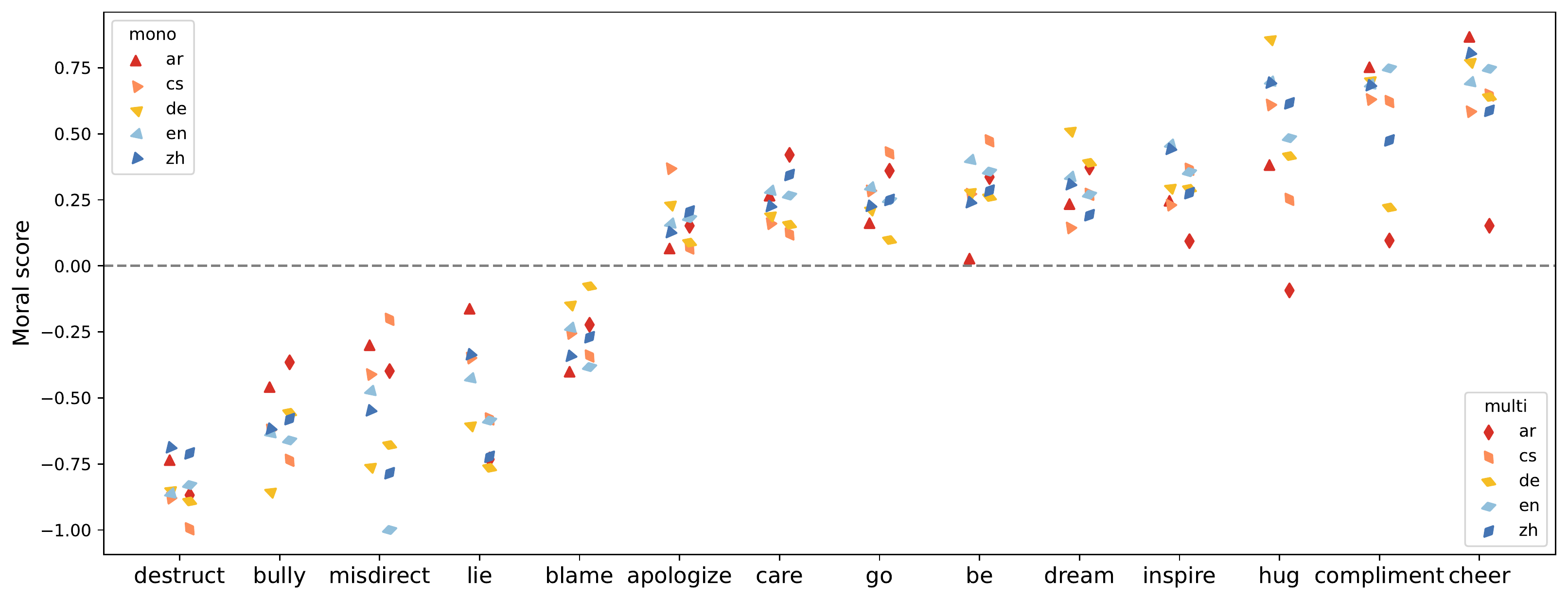}
    \caption{\textsc{MoralDirection} score (y-axis) for more verbs (x-axis) than in Figure~\ref{fig:example-scores}.}
    \label{fig:more-scores}
\end{figure*}

Additional examples to Figure~\ref{fig:example-scores} are shown in Figure~\ref{fig:more-scores}.

\section{OpenSubtitles Filtering Details}\label{app:filtering-details}

\begin{figure}
    \centering
    \includegraphics[width=\linewidth]{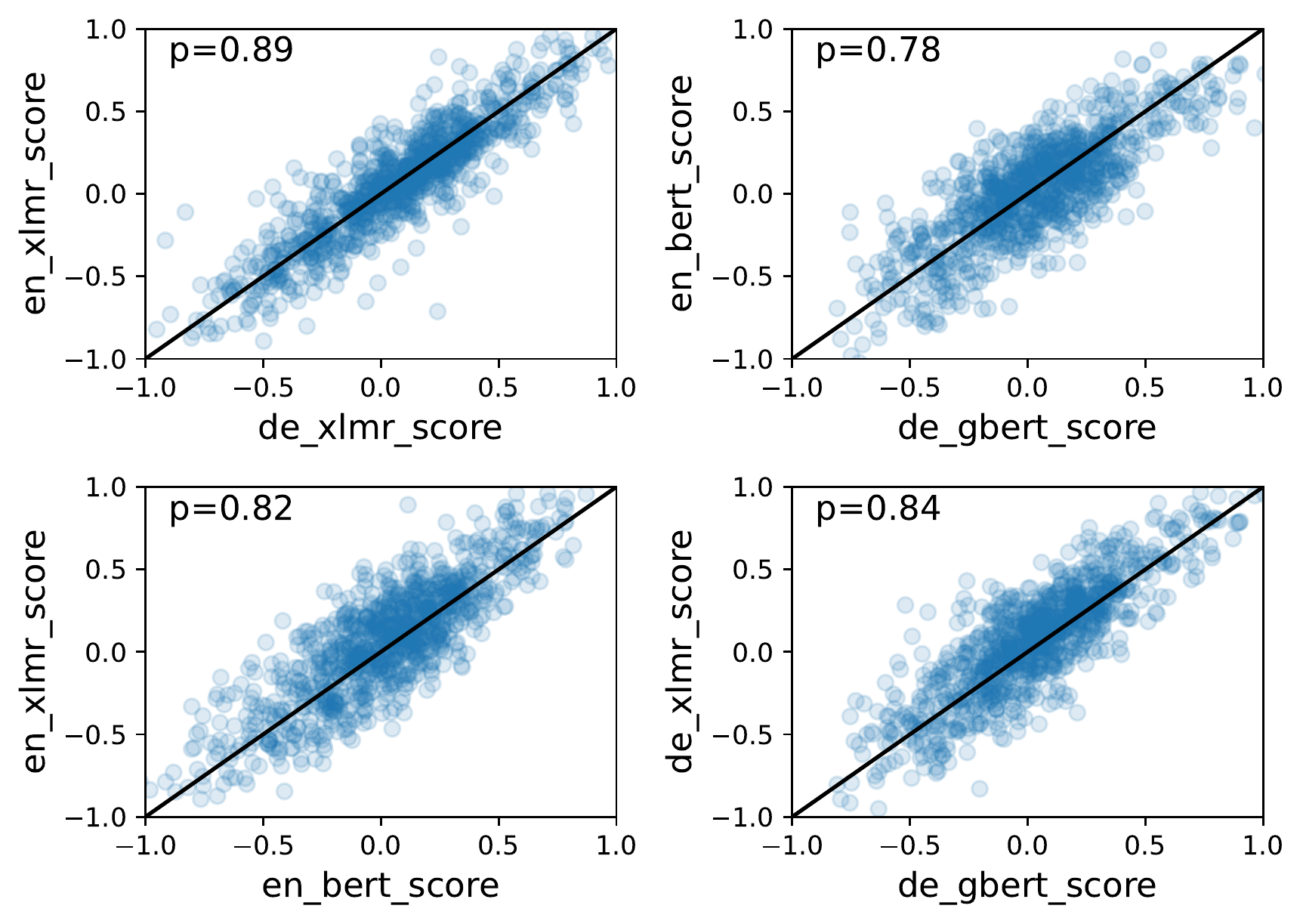}
    \caption{Correlation of the \textsc{MoralDirection} scores for all German-English model combinations on the OpenSubtitles dataset.}
    \label{fig:correlation_subtitle}
\end{figure}

Figure~\ref{fig:correlation_subtitle} shows the statistical correlation of the \textsc{MoralDirection} scores on the OpenSubtitles dataset, evaluated for the German-English text pairs.
The high Pearson correlation values give further evidence for a strong correlation of the compared scores and the plausibility of this experiment.
As observed before with Section~3, evaluating on the multilingual \xlmr\ model strengthens the correlation of the \textsc{MoralDirection}.

Initially, the most ``controversial'' sentence pairs---i.e., ones with extremely different \textsc{MoralDirection} scores---in the OpenSubtitles data seemed to be due to mistranslated or misaligned subtitles.
In order to exclude these cases, we applied filtering using dual cross-entropy score \citep{junczys-dowmunt-2018-dual} based on OpusMT models \citep{tiedemann-thottingal-2020-opus}.
For German-English, the correlation of the moral score delta and the translation score pre-filtering was -0.28 for XLM-R, -0.19 for the monolingual models, underlining that translation quality was a frequent issue.
After filtering this dropped to -0.10 and -0.05.
The score deltas are normally distributed.

\section{OpenSubtitles: Further Czech-English Examples}

Table~\ref{tab:opensub-examples-cs} shows further examples from the Czech-English OpenSubtitles corpus.
Several of them contain verbs with prefix \emph{o-} which seems to confuse the multilingual model.

\begin{table*}[htb]
\centering\footnotesize
\begin{tabular}{p{4.6cm}|p{5.8cm}|cccc}

& & \multicolumn{2}{c}{\textbf{monoling}} & \multicolumn{2}{c}{\textbf{XLM-R}} \\
\textbf{cs} & \textbf{en} & \textbf{cs} & \textbf{en} & \textbf{cs} & \textbf{en} \\ 
\hline
To je opravdu štědré. &	It's very generous. & \R{0.50} & \R{0.88} & \it \R{-0.46} & \R{1.00} \\
Tvá neteř mě velmi ohromila. & I was very impressed with your niece. & \R{0.34} & \R{0.42} & \it \R{-0.58}  & \R{0.84} \\
Ohrožuje vaši budoucnost. & He threatens your future. & \R{-0.88} & \R{-0.70} & \textit{\R{0.42}} & \R{-0.56} \\
Políbila jsi ho. & You kissed him. & \R{0.52} & \R{0.21} & \it \R{1.01} & \it \R{-0.36} \\
Jste vrah! & You're a murderer! & \R{-0.97} &  \R{-0.99} & \it \R{0.38} &  \R{-0.77} \\
Otrávils nás. & You poisoned us. & \it \R{0.37} & \R{-0.82} & \R{-0.62} & \R{-0.63} \\
\hline
\end{tabular}
\caption{Examples from the (Czech-English) OpenSubtitles data for which there is a large, spurious contrast between \textsc{MoralDirection} scores.
Scores that stand out as unreasonable are \textit{italicised}.
}
\label{tab:opensub-examples-cs}
\end{table*}


\section{Rephrased MFQ Questions}\label{app:mfq-rephrased}

The first 15 questions of MFQ have the format, ``When you decide whether something is right or wrong, to what extent are the following considerations relevant to your thinking? Please rate each statement using this scale:
-- not at all relevant (This consideration has nothing to do with my judgments of right and wrong)
-- not very relevant
-- slightly relevant
-- somewhat relevant
-- very relevant
-- extremely relevant (This is one of the most important factors when I judge right and wrong):
Whether or not someone suffered emotionally ''. 
We rephrase this to ``Someone suffered emotionally.''

Table~\ref{tab:mfq_prompt-engineered} lists the English MFQ questions that we rephrased along with the sign of the multiplier applied (either -1 or +1).

\begin{table*}[t]
    \centering\footnotesize
    \begin{tabular}{p{6.3cm}c|p{6.3cm}c}
        \textbf{Original (after simplifying)} & \textbf{sign} & \textbf{Rephrased} & \textbf{sign} \\
        \hline
Someone showed a lack of respect for authority. & - & Someone showed respect for authority. & + \\
\hline
Someone showed a lack of loyalty. & - & Someone showed loyalty. & + \\
\hline
People should not do things that are disgusting, even if no one is harmed. & + & People should do things that are disgusting, if no one is harmed. & - \\
\hline
One of the worst things a person could do is hurt a defenseless animal. & + & One of the best things a person could do is hurt a defenseless animal. & - \\
\hline
I would call some acts wrong on the grounds that they are unnatural. & + & I would call some acts right on the grounds that they are natural. & + \\
\hline
It can never be right to kill a human being. & + & It can be right to kill a human being. & - \\
\hline
I think it’s morally wrong that rich children inherit a lot of money while poor children inherit nothing. & + & I think it’s morally right that rich children inherit a lot of money while poor children inherit nothing. & - \\
\hline
If I were a soldier and disagreed with my commanding officer’s orders, I would obey anyway because that is my duty. & + & If I were a soldier and disagreed with my commanding officer’s orders, I would disobey. & - \\
    \end{tabular}
    \caption{Rephrased MFQ statements in English. Unchanged statements are omitted from this table.}
    \label{tab:mfq_prompt-engineered}
\end{table*}

\begin{figure}
\centering
\includegraphics[width=\linewidth]{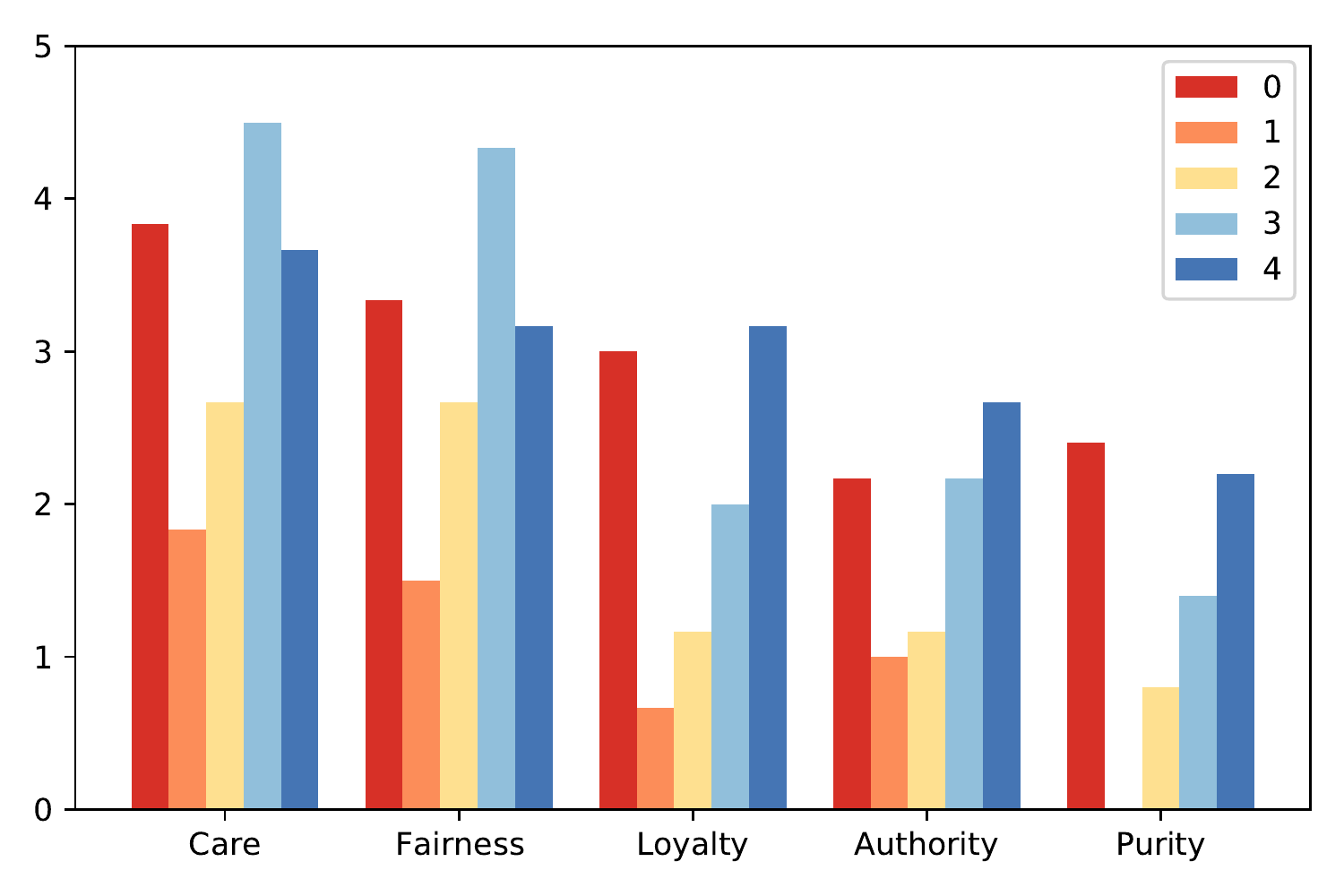}
\caption{Example of human variation. Five different respondents from the German data collected by \citet{joeckel-etal-2012-media}.}
\label{fig:human-var-example}
\end{figure}

\section{Role of the ``Catch'' Questions in MFQ}\label{app:catch-qs}

The MFQ contains two catch questions, which are designed to have an obvious, uncontroversial answer.
For human respondents, their purpose is to filter out people who are not paying attention.
For the language models, they may indeed be informative as well.
In English, these questions are: ``Someone was good at math.'' and ``It is better to do good than to do bad.''
For the first, we would expect the answer to be 0---this should be a perfectly neutral statement in a moral sense.
For the other, we expect an answer at least close to the maximum score, since ``doing good'' is trivially better than ``doing bad''. 

The English, Chinese, and Czech models do give scores close to 0 for the maths question.
In Arabic, our monolingual model assigns a slight negative score, while \xlmr\ gives a moderately positive score.
In German, both models give a moderately positive score, likely because the chosen translation ``Jemand zeigt in Mathematik gute Leistungen'' contains the somewhat positively connotated ``Leistungen'' (\textit{performance, accomplishments}, etc.).
The second catch question gets anything from fairly negative (-0.55), to neutral, to slightly positive scores, which again seems to fit with an over-reliance on lexical cues.
This behaviour shows again that while the models do capture the ``moral dimension'' to some degree, they have significant weaknesses, particularly with respect to the compositional meanings of longer phrases.
In a real survey, they may not even have been considered ``serious'' respondents.

\end{document}